\documentclass[letterpaper, 10 pt, conference]{ieeeconf}  % Comment this line out if you need a4paper

\IEEEoverridecommandlockouts                              % This command is only needed if 
\usepackage{graphicx}
\usepackage{amsmath} % assumes amsmath package installed
\usepackage{amssymb}  % assumes amsmath package installed
\usepackage{xcolor}
\usepackage{float}
\usepackage{multirow}
\usepackage[stable]{footmisc}
\usepackage[utf8x]{inputenc}
\usepackage{hyperref}

\title{\LARGE \bf
Finding Locomanipulation Plans Quickly \\ in the Locomotion Constrained Manifold
}

\author{Steven Jens Jorgensen$^{1,2}$, Mihir Vedantam$^{3}$, Ryan Gupta$^{3}$, Henry Cappel$^{3}$, and Luis Sentis$^{3}$% <-this % stops a space
\thanks{$^{1}$The author is supported by a NASA Space Technology Research Fellowship (NSTRF).}%
\thanks{The authors are with the $^{2}$Department of Mechanical Engineering and $^{3}$Department of Aerospace Engineering in the University of Texas at Austin}%
}

\begin{document}

\maketitle
\thispagestyle{empty}
\pagestyle{empty}

%%%%%%%%%%%%%%%%%%%%%%%%%%%%%%%%%%%%%%%%%%%%%%%%%%%%%%%%%%%%%%%%%%%%%%%%%%%%%%%%
\begin{abstract}
We present a method that finds locomanipulation plans that perform simultaneous locomotion and manipulation of objects for a desired end-effector trajectory. Key to our approach is to consider a generic locomotion constraint manifold that defines the locomotion scheme of the robot and then using this constraint manifold to search for admissible manipulation trajectories. The problem is formulated as a weighted-A* graph search whose planner output is a sequence of contact transitions and a path progression trajectory to construct the whole-body kinodynamic locomanipulation plan. We also provide a method for computing, visualizing and learning the locomanipulability region, which is used to efficiently evaluate the edge transition feasibility during the graph search. Experiments are performed on the NASA Valkyrie robot platform that utilizes a dynamic locomotion approach, called the divergent-component-of-motion (DCM), on two example locomanipulation scenarios. 

% use path progression trajectory next time
% IHMC acronym not defined
% add A* approach on conclusion
% goal vertex is assumed unreachable

%This is done by considering that the locomotion approach for the robot is given creating a locomotion manifold for which a search on admissible manipulation trajectories  We present a method to compute, visualize, and learn locomanipulability regions We find locomanipulation plans in the manifold of the locomotion scheme. 

\end{abstract}

%%%%%%%%%%%%%%%%%%%%%%%%%%%%%%%%%%%%%%%%%%%%%%%%%%%%%%%%%%%%%%%%%%%%%%%%%%%%%%%%
\section{INTRODUCTION}
To exploit the full capabilities of humanoid robots in human-centered environments, it is critical that the robots are able to efficiently interact with objects designed for human use. However, much of the success with locomanipulation of objects has been seen with wheeled-based mobile-manipulators \cite{meeussen2010autonomous,ruhr2012generalized,welschehold2017learning, arduengo2019versatile}. For instance, \cite{arduengo2019versatile} shows robust manipulation of kinematically constrained objects such as doors and cabinets. The success of wheeled-bases is unsurprising as the manifold for locomotion and manipulation is continuous which simplifies the search for feasible plans. However, robots with limbs rely on contact transitions to perform locomotion. As breaking and making contacts are discrete decisions that introduce discontinuity and can even be combinatorial when finding an appropriate contact mode schedule \cite{posa2013direct}, it is non-trivial to identify a sequence of dynamically feasible contact transitions during manipulation. 

One way to address the discontinuity issue with coupled locomotion and manipulation of limbed robots is to treat the floating degrees of freedom of the robot to be controllable, for instance by constraining it to SE(2), then solving the locomanipulation problem as one would with a wheeled-base robot and finding a satisfying quasi-static sequence of footsteps \cite{dalibard2010manipulation}. A more recent approach treats the end-to-end locomanipulation problem as rearrangement planning, however it also only outputs quasi-static solutions \cite{mirabel2016constraint}. The difficulty of handling contact transitions while performing manipulation is the reason that whole-body manipulation of objects by limbed robots are often performed by maintaining the same stance configuration throughout the entire manipulation trajectory. For example, in \cite{berenson2011task}, bi-manual manipulation of a humanoid robot is performed with the same stance configuration. In \cite{ferrari2017humanoid}, locomotion, locomanipulation, and manipulation zones are constructed to approach the object in the manipulation zone and perform the manipulation task with a fixed stance.
Furthermore, all the previously mentioned approaches are only able to output quasi-static solutions.

\begin{figure}
\centerline{\includegraphics[width=0.75\columnwidth]{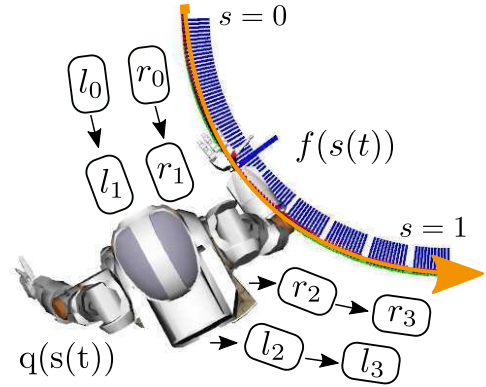}}
\caption{A top-view visualization of the considered locomanipulation problem definition. Given a manipulation constraint end-effector path/s described by $f(s)$, the goal is to find a progression trajectory $s(t) \in [0,1]$ and a sequence of contact transitions $(l_i,r_i)$ such that the resulting whole-body trajectory $q(s(t))$ also satisfies the prescribed locomotion manifold. A solution is a feasible locomanipulation plan.}
\label{fig:problem_definition}
\end{figure}

In contrast, we present an approach that is able to find dynamic locomanipulation plans with kinodynamic whole-body solutions. This is done by first defining the locomotion constraint manifold and then finding manipulation plans that satisfy the original locomotion constraint. This is equivalent to finding manipulation trajectories in the \textit{nullspace} of the locomotion. As a motivating example, we consider the locomotion constraint manifold to be the task-space trajectories generated by the  dynamic locomotion approach called the divergent-component-of motion (DCM) \cite{englsberger2015three} that is used on the NASA Valkyrie robot \cite{radford2015valkyrie} with a momentum-based whole-body controller \cite{koolen2016design}. A benefit of our approach is that kinodynamic trajectories are automatically produced by virtue of selecting a dynamic locomotion scheme. Additionally, if the locomotion approach has stability properties, the resulting whole-body trajectories will also have these properties. Note again that our approach is invariant to the locomotion scheme and the underlying whole-body controller. 

Next, we formulate locomanipulation as the following problem. Given  SE(3) end-effector trajectories for the hands, the goal is to find a progression trajectory for the hands with a satisfying sequence of footsteps such that the resulting whole-body trajectory also satisfies the locomotion constraint manifold (See Fig ~\ref{fig:problem_definition}). We solve this as a graph search problem with a weighted A* as the planner. To efficiently compute feasible edge transitions that can be manipulation, locomotion, or locomanipulation trajectories we introduce a method for learning the locomanipulability regions of the robot with the prescribed locomotion constraint manifold with a neural-network based classifier. The solution of the planner is a kinematically feasible trajectory that respects joint limits. Kinodynamic satisfiability is also achieved if the external disturbance of the manipulation task can be sufficiently rejected or compensated by the low-level whole-body controller. Finally, we show that we are able to generate fast locomanipulation plans on two toy example problems.

Our paper has two key contributions. First, we introduce a novel method to compute, visualize, and learn the locomanipulability regions, defined as the region in which both manipulation and locomotion are possible. Second, we introduce a fast weighted A* planner formulation which uses the learned locomanipulation regions to find satisfying locomanipulation plans.

\subsection{Related Works on Locomanipulation}
While the problem of finding locomanipulation plans is discussed here, there are other recent works on locomanipulation-related problems such as \cite{borras2015whole, farnioli2016toward, kaiser2016towards, asfour2018dualities}. In \cite{borras2015whole}, a taxonomy of locomanipulation poses is presented as well as an example analysis of required pose transitions to climb stairs. \cite{kaiser2016towards} provides a ground work for understanding environment affordances for locomanipulation. \cite{asfour2018dualities} extends \cite{borras2015whole} and \cite{kaiser2016towards} by using data to auto-generate a pose transition graph and testing their affordance classifications on a mobile manipulator with a wheeled base. 

 We previously described existing quasi-static approaches that used search based algorithms to solve locomanipulation problems. However, our idea of dynamic locomanipulation by finding manipulation trajectories in the nullspace of locomotion has been previously pursued in \cite{settimi2016motion}. In their work, primitives for both locomotion and manipulation are generated beforehand. Then, an offline RRT-based planner is used to find locomanipulation plans in the intersection of the primitives' image spaces. Our work differs from them in a few ways. First, we have a different problem and planner formulation for finding locomanipulation plans. For instance, we consider manipulating objects with predefined manipulation trajectories (e.g. as described by affordance templates (ATs) \cite{hart2014affordance}). Next, because their method consists of a search over the null space of the prioritized motion primitive, pure locomotion or pure manipulation phases are not considered in their framework, which is not a limitation in our planner. Another work, \cite{bouyarmane2012humanoid} uses a search based algorithm for planning contact transitions for the purposes of locomotion and manipulation for many types of robots. However, the coupled locomotion and manipulation problem are not considered. More recently, \cite{bouyarmane2018non} presents a method for addressing the coupled locomotion and manipulation problem as we do here. However, their results are on low-dimensional degree-of-freedom systems with no consideration of joint limits. A complete kinodynamic planner utilizing SQP methods was presented in \cite{dai2014whole}, but it is prohibitively expensive to compute and requires good initial conditions.

\section{Approach Overview}
To find locomanipulation plans, the key idea is to first consider that the locomotion scheme for the robot is provided ahead of time. This constrains the possible locomotion trajectories that the robot can execute. Then, locomanipulation is achieved by finding admissible manipulation trajectories that satisfy both the original locomotion constraint and the desired manipulation end-effector trajectory. We consider limbed robots of humanoid form, but the ideas presented here can also work with other multi-limbed robots.

%There are two argued benefits of this approach. First, we are able to generate kinematic manipulation plans that utilize the existing locomotion capability of the robot. Second, the locomotion approaches of existing robots are often based on well-known low-order dynamics such as the linear inverted pendulum (LIP) with known stability properties. Thus, when the kinematics trajectory, $q(t)$, is extracted from the task space trajectories, we have the added benefit of having locomotion capabilities with the same properties.

\begin{figure*}
\centerline{\includegraphics[width=1.8\columnwidth]{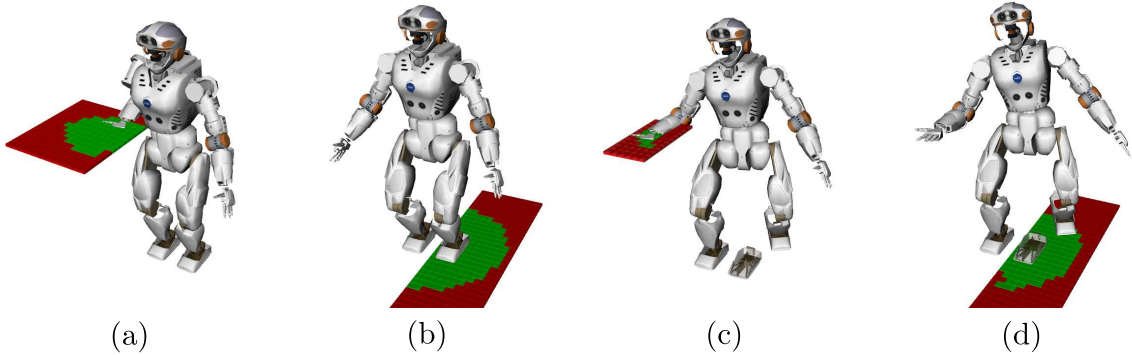}}
\caption{(a) A visualization of the manipulation reachability of the right hand, (b) the locomotion contact transition reachability, (c) the locomanipulation region in the end-effector space of the hand, and (d) the locomanipulation region in the contact transition space. The transparent left foot in (c) and (d) indicates the starting stance. Notice that the locomanipulability regions are always a subset of the reachability regions for both manipulation and locomotion.}
\label{fig:locomanipulability region}
\end{figure*}

\subsection{Problem Definition}
The locomanipulation problem is formulated as follows: given a desired end-effector path trajectory $f(s)$ with $s \in [0,1]$, the goal is to find a manipulation progression variable trajectory $s(t)$ and a footstep sequence trajectory such that the resulting whole-body trajectory $q(s(t))$ satisfies the desired end-effector path trajectory $f(s)$ and the locomotion constraint manifold. For instance, suppose the robot's task is to open a door  (See Figure \ref{fig:problem_definition}). The desired end-effector trajectory for the hand can be defined in terms of the trajectory of the handle as the door opens. This is similar to how ATs \cite{hart2014affordance} or task space regions (TSRs) \cite{berenson2011task} would define the robot interface to the door. At any point in time, the robot may decide to pull on the door, take a footstep, or do both at the same time. An action which pulls the door is a progression of the $s$ variable from $s_i$ to $s_{i+1}$. We call this an increment of the manipulation variable by some $\Delta s$. 

\subsection{Defining the Locomotion Constraint Manifold}
Existing locomotion schemes in limbed robots for example are performed with quasi-static, capture-point, divergent component of motion (DCM) \cite{englsberger2015three}, time-velocity-reversal (TVR) \cite{kim2018control}, or centroidal-momentum based planners \cite{ponton2018time}. These high-level planners output CoM trajectories (and sometimes momentum trajectories) for a given sequence of contact modes. Consequently, to satisfy these centroidal trajectories with contact constraints, task space trajectories for the end-effectors such as the feet, palm, pelvis, etc, also have to be constructed by an accompanying planner. Additionally, these high-level planners which constitute the locomotion scheme are typically injective. That is, for a given sequence of contact modes and an initial condition of the robot configuration $q$, $\dot{q}$, it will always output the same task space trajectories, $x(t)$ for the CoM and end-effectors. For humanoid walking these task space trajectories could be
\begin{align}
    x_{\textrm{L}}(t) = [x_{\textrm{COM}}(t), x^{\textrm{left}}_{\textrm{foot}}(t),x^{\textrm{right}}_{\textrm{foot}}(t), x_{\textrm{pelvis}}(t)]^T,
\end{align}
with the task spaces defined such that $x_{\textrm{COM}} \in \mathbf{R}^3$, $x_{\textrm{foot}}(t) \in SE(3)$, and $x_{\textrm{pelvis}} \in SO(3)$. Additionally these tasks will have corresponding locomotion task Jacobian,
\begin{align}
\label{eq:locomotion_task_jacobian}
    \Delta x_{\textrm{L}}(t) = J_{\textrm{L}}(q(t)) \Delta q(t) 
\end{align}

Thus, the locomotion scheme provides a constraint manifold, Eq.~\ref{eq:locomotion_task_jacobian}, that needs to be satisfied when finding admissible manipulation plans.

\subsection{Defining the Locomanipulability Region}
We define the locomanipulability region to be the area in which both locomotion and manipulation tasks are feasible. By constraining the locomotion scheme, we are able to test whether a particular manipulation trajectory (e.g. a hand end-effector trajectory) satisfies a given locomotion manifold. Equivalently, a manipulation constraint can be initially set and used to check whether the original locomotion plan is still valid. When both manipulation and locomotion trajectories are feasible, locomanipulation becomes possible. From the problem definition, the manipulation constraint can be described in terms of s, namely:
\begin{align}
    \Delta x_{\textrm{M}}(s) = J_{\textrm{M}}(q(s)) \Delta q(s),
\end{align}
where the subscript $M$ indicates manipulation tasks in SE(3) with its corresponding Jacobian.

Numerically checking whether a manipulation trajectory $x_{\rm M} (s)$is admissible for a given locomotion manifold $x_{\rm L}(t)$ is checked with a series of inverse-kinematics (IK) that simulate the whole-body controller on the robot (See Sec.~\ref{sec:ik_config_trajectory}).

Similar to reachability regions  \cite{zacharias2007capturing} for manipulation (Fig.~\ref{fig:locomanipulability region}a) and locomotion (Fig.~\ref{fig:locomanipulability region}b), we can define the locomanipulability region as a region in space for which both locomotion and manipulation tasks are possible. This region can be defined either in the end-effector space (Fig.~\ref{fig:locomanipulability region}c) or the contact transition space (Fig.~\ref{fig:locomanipulability region}d). For the former, if the contact transition is fixed (ie: the robot is set to take a left footstep), there will only be a small region in the end-effector space for which manipulation trajectories are possible. For the latter, suppose the robot's right hand is to be constrained in a particular pose in $SE(3)$, then the region on the floor for which footstep transitions are possible will be the locomanipulation region defined in the contact transition space. Fig.~\ref{fig:locomanipulability region}d).  

\section[IMPLEMENTATION DETAILS]{IMPLEMENTATION DETAILS}
In addition to the following sections, our software implementation is also available. \footnote{\href{https://github.com/stevenjj/icra2020locomanipulation}{https://github.com/stevenjj/icra2020locomanipulation}}
%The NASA Valkyrie robot uses DCM based trajectories which we will use to define our constraint manifold.

\subsection{Locomotion Manifold Parameters}
The following task space trajectories for CoM, feet and pelvis are based on a simplified behavior of IHMC's\footnote{The Institute for Human Machine \& Cognition} walking controller on NASA's Valkyrie robot. For a given foot contact sequence and initial condition of the COM state, the DCM generates a COM trajectory based on a specified swing foot time, double support transfer time, and final settling time. At the beginning and end of the DCM trajectory, the desired virtual repellant point (VRP) is set at the support polygon center, so that the beginning and ending of each walking trajectory will have the CoM at the support polygon center. 
In addition to the CoM trajectory, satisfying task space trajectories for the feet and pelvis still need to be set. Throughout the walking trajectory the pelvis orientation is always the average of the orientation of the feet using spherical linear interpolation (SLERP)\cite{shoemake1985animating}. The average of the feet orientation and position is referred to as the midfeet frame.
\begin{align}
    x_{\textrm{pelvis}}(t) = \textrm{SLERP}(0.5, x^{\rm left}_{\rm foot}, x^{\rm right}_{\rm foot}).
\end{align}
If at the start of the DCM trajectory the pelvis orientation is not equal to midfeet frame orientation due to manipulation tasks, a hermite quaternion curve \cite{kim1995general} is used to interpolate the pelvis orientation before the robot begins to walk. %Note that the pelvis position is implicitly controlled by the CoM so it is not controlled.  

For the swing foot position, We use two hermite curves with boundary conditions at the apex of the foot swing. At the apex of the swing, the velocity of the foot is set to be the average velocity of the swing foot defined as
\begin{align}
\dot{x}_{\rm foot}(\frac{t_{\rm swing}}{2}) = \frac{\Delta x_{\rm foot}}{t_{\rm swing}},
\end{align}
where $\Delta x_{\rm foot}$ is the total distance traveled by the swing foot and $t_{\rm swing}$ is the swing time. The swing foot orientation is constructed with a single hermite quaternion curve with zero angular velocity boundary conditions. Finally, if the foot is in stance or in double support, its position and orientation are held constant.

\subsection{IK Configuration Trajectory}
\label{sec:ik_config_trajectory}
For a given desired locomotion and manipulation task space trajectories, a feasible IK trajectory with these two tasks simultaneously implies that the desired locomanipulation trajectories are feasible. For a given footstep contact sequence, we obtain a locomotion task space trajectory $x_{\rm L}(t)$ with duration $\Delta T$. Similarly, for a given increment of the manipulation variable, $\Delta s$, we obtain a manipulation task trajectory $x_{\rm M}(s)$. The locomotion and manipulation trajectories can be parameterized by an indexing variable $i \in \{0, 1, ..., N \}$, a discretization factor $N$, and making the following substitutions
\begin{align}
    t(i) = t_o + \frac{i \Delta T}{N}, \\
    s(i) = s_o + \frac{i \Delta s}{N},
\end{align}
with $t_o$ and $s_o$ the initial values of $t$ and $s$ at $i = 0$. We can then create the locomanipulation task by stacking the tasks and their Jacobians with $x_{\rm LM}(i) = [x_{\rm L}^T(i), x_{\rm M}^T(i)]^T$ and $J_{\rm LM}(i) = [J^T_L(i), J^T_M(i)]^T$. We also add a posture joint position task $J_{P}$ with task errors $\Delta x_{P}$ in the the torso which helps condition the trajectories to be near a deisred nominal pose. Then, the IK configuration trajectory, which mirrors the controller behavior of the robot, is performed using the following equations.
\begin{align}
    \label{eq:lm_task_error}
    & \Delta x_{\rm LM}(i) = x_{\rm LM}(i) - x_{\rm LM}(q(i)), \\ 
    \label{eq:ik_dq}
    & \Delta q(i) = k_p \cdot \overline{J}_{\rm LM}(i) \Delta x_{\rm LM}(i), + \overline{( J_{P} N_{\rm LM})}(\Delta x_{P}) \\
    \label{eq:ik_q_update}
    & q(i+1) = c( q(i) + \Delta q(i) )
\end{align}
where $\overline{X} = (A^{-1}X^T)(X A^{-1} X^T)^{\dagger}$ is the dynamically consistent pseudoinverse with $A$ being the inertia matrix for a robot configuration $q(i)$ and $^\dagger$ indicates the pseudoinverse. $N_{\rm LM} = (I - \overline{J_{\rm LM}} J_{\rm LM})$ is the nullspace of the locomanipulation task, with $I$ the identity matrix. The task error at the $i$-th index is defined by Eq.~\ref{eq:lm_task_error} in which $x_{\rm LM}(q(i))$ is the current task space poses given the robot configuration. The configuration change is obtained using Eq.~\ref{eq:ik_dq} with $k_p$ a scalar gain, and a configuration update is performed with Eq.~\ref{eq:ik_q_update} with $c(\cdot)$ being a clamping function that ensures joint limits are not exceeded. Finally, Eqs.~\ref{eq:lm_task_error}-\ref{eq:ik_q_update} are iteratively repeated.  If an iteration causes the task error to increase, backtracking on $k_p$ is performed by updating it with $k^*_p = \beta k_p$ with $\beta = 0.8$. The trajectory converges when all the $\Delta x_{LM}(i)$ are driven to 0. The trajectory fails to converge when the norm of $\Delta q(i)$ goes below 1e-12.
%Note that this IK trajectory scheme was selected since the typical QP-based wholebody low-level controllers perform this operation

\begin{table}
\caption{ Classifier feature vector }
\centering
\label{table:classifier_feature_vector}
\begin{tabular}{|c||c|c|}
\hline
\textbf{Type}  & \textbf{Feature Name} & \textbf{Dim} \\
\hline
$\mathbf{R}^1$ & Stance Leg & 1 \\
\hline
$\mathbf{R}^1$ & Manipulation Type & 1 \\
\hline
$SE(3)$ & Pelvis Starting Pose & 6 \\
\hline
$SE(3)$ & Swing Start and Land Foot Pose & 12 \\
\hline
$SE(3)$ & Left and Right Hand Poses & 12 \\
\hline
\end{tabular}
\end{table}

\subsection{Learning the Locomanipulability Region}
\label{sec:learning-locomanipulability}
%The NASA Valkyrie robot uses IHMC's footstep planner which defines a kinematic reachability region in terms of the footstep landing location with respect to the stance foot. 
When deciding whether or not a contact transition and a progression variable $\Delta s$ change is possible, instead of running the full IK trajectory to check for convergence, we instead learn a classifier that learns the result of the IK trajectory for the given task space inputs.  Similar to the approach presented in \cite{lin2019efficient} that used a neural network for classifying contact transition feasibility, the classifier used here will learn the trajectory feasibility but instead with a manipulation constraint. The classifier is a 3-layer fully connected network with 100 ReLu units per layer \cite{nair2010rectified} and a sigmoid activation function for binary classification. The network is trained with the keras framework \cite{chollet2015}.

\begin{table}
\caption{ Edge Feasibility Check Performance }
\centering
\label{table:edge_feasibility_check_performance}
\begin{tabular}{|c||c|}
\hline
\textbf{Transition feasibility check type}  & \textbf{Time per edge (seconds)} \\
\hline
IK Trajectory & (2.11 $\pm$ 0.13)s \\
\hline
Neural Network Classifier & (1.44 $\pm$ 0.18) $\cdot 10^{-3}$ s \\
\hline
\end{tabular}
\end{table}

The input vector, $p(v_1, v_2; s)$, used for the neural network classifier can be seen in Table \ref{table:classifier_feature_vector}. The input vector is a function of the two graph vertices $(v_1, v_2)$ as described in Sec. \ref{sec:A*}, but it is parameterized by the location of the end-effector along the manipulation trajectory, $f(s)$. The stance leg is a binary variable that indicates which leg is the stance leg (left, right). Similarly, the manipulation type indicates the manipulator end-effectors (left, right, or both hands). The remainder of the features are the 6D poses of the specified robot body part with respect to the stance foot. As there are two choices for the swing leg and three choices for manipulation type, there are six possible contact transitions to consider. For each contact transition type, the training data is generated by randomly generating the upper body joint configurations, and randomly selecting a foot landing location w.r.t to the stance foot as the origin. The pelvis pose is also randomly generated in the convex hull of the feet. For a particular manipulation type, we fix the manipulator pose and solve a series of IKs (Sec.~\ref{sec:ik_config_trajectory}) that simulate the robot's whole body controller to check if the locomanipulation trajectory is feasible.

The output of the classifier is a prediction score, $y(\cdot) \in [0, \ 1]$, that indicates the feasibility of the queried transition. Since the classifier is only trained on data that represents locomanipulation with a fix manipulator pose ($\Delta s = 0)$, additional steps are taken to use the classifier for manipulation-only decisions and locomanipulation decisions with a moving manipulator pose ($\Delta s \neq 0)$. When considering the manipulation only case, the manipulation trajectory is discretized into $N_m$ equidistant points and a step in place trajectory is queried from the neural network for each point. This method assumes that if the discretized points are in the locomanipulation region then the entire trajectory must be as well. The lowest score is then taken as the feasibility score. For the locomanipulation with a moving manipulator pose case ($\Delta s \neq 0$), a similar discretization is used but instead of testing a step in place, the specified swing foot trajectory is tested at each of the points. Once again, the lowest score is taken as the feasibility score. A succinct description for the feasibility score is written as
    \begin{align}
        n(v_1, v_2) = 
        \begin{cases}
            y(p(v_1, v_2; s)) \textrm{,} & \Delta s = 0 \\
            \min\limits_{i= 1, \dots, N_m} y(p(v_1,v_2 ; s_i)) \textrm{,} & \Delta s \neq 0.
        \end{cases}
        \label{eq:feasFunc}
    \end{align}
\subsection{Weighted A* Formulation} \label{sec:A*}
Finding locomanipulation plans is formulated as a low-dimensional graph search problem, $G = (V, E)$. Each vertex $v \in V$ is a locomanipulation state $v = (s, x_{\rm feet}, y_{\rm feet}, \theta_{\rm feet}) \in \mathbf{R}^7$, where $s$ is the manipulation variable state, and $(\cdot)_{\rm feet}$ are the states of the left and right feet. The states are discretized from the starting position of the robot. We assume that the starting position of the robot with $f(s=0)$ is such that the configuration is in the locomanipulation region. Only a finitely sized lattice is considered by defining a kinematic reachability limit from a certain radius (e.g. 1.5m) from $f(s)$. An edge $e \in E$ in the graph is a transition between two vertices $v_1$ and $v_2$ which can have a $\Delta s$ change that progresses the manipulation variable, and/or a foot contact transition. This enables the planner to make a decision between performing manipulation, locomotion, or locomanipulation trajectories. 
\subsubsection{Edge Cost}
A contact transition between two vertices has the following edge transition cost.
\begin{align}
\label{eq:edge_cost}
\Delta g(v_1, v_2) = &w_s \cdot (1-s) + w_{\rm step} + \\ 
                     & w_{\rm L} \cdot r(v_2) + w_d \cdot (1 - n(v_1, v_2)), \nonumber
\end{align}
where $w_s$ encourages the progression of the manipulation trajectory, $w_{\rm step}$ is a scalar cost of taking a footstep, $w_L$ penalizes states that deviate from a suggested body path $r(v_2)$, and $w_d$ penalizes edge transitions that have low feasibility computed by the feasibility score $n(v_1, v_2)$. 

The suggested body path can be an output from the same high-level planner that produced the end-effector trajectory for $f(s)$. Here we first compute $T^{{s_0}}_{\rm foot}$, which is the fixed transform between the initial end-effector pose, $f(s=0)$, and the starting stance foot pose. For a given $s$, we then transform the initial stance to the corresponding pose of $f(s)$ using $T^{{s_0}}_{\rm foot}$. Then $r(v_2)$ is computed as the norm of the difference between a foot landing location in $v_2$ and the aforementioned transformation.

Since the planner is successful when it finds a feasible path to a state such that $s = 1$, notice that maximizing for feasibility is not necessarily the best course of action as the planner can mindlessly perform contact transitions that are feasible. %Depending on the classifier performance, it is also possible for the classifier to make mistakes, so overly trusting the classifier can lead to non-convergence. 
Therefore, a trade-off has to be performed between progressing the manipulation variable,  $s$, attempting a transition using the suggested body path $r(v_2)$,  deciding whether or not to make a footstep transition at all, or choosing a vertex that maximizes for feasibility.

\begin{figure*}
\centerline{\includegraphics[width=1.85\columnwidth]{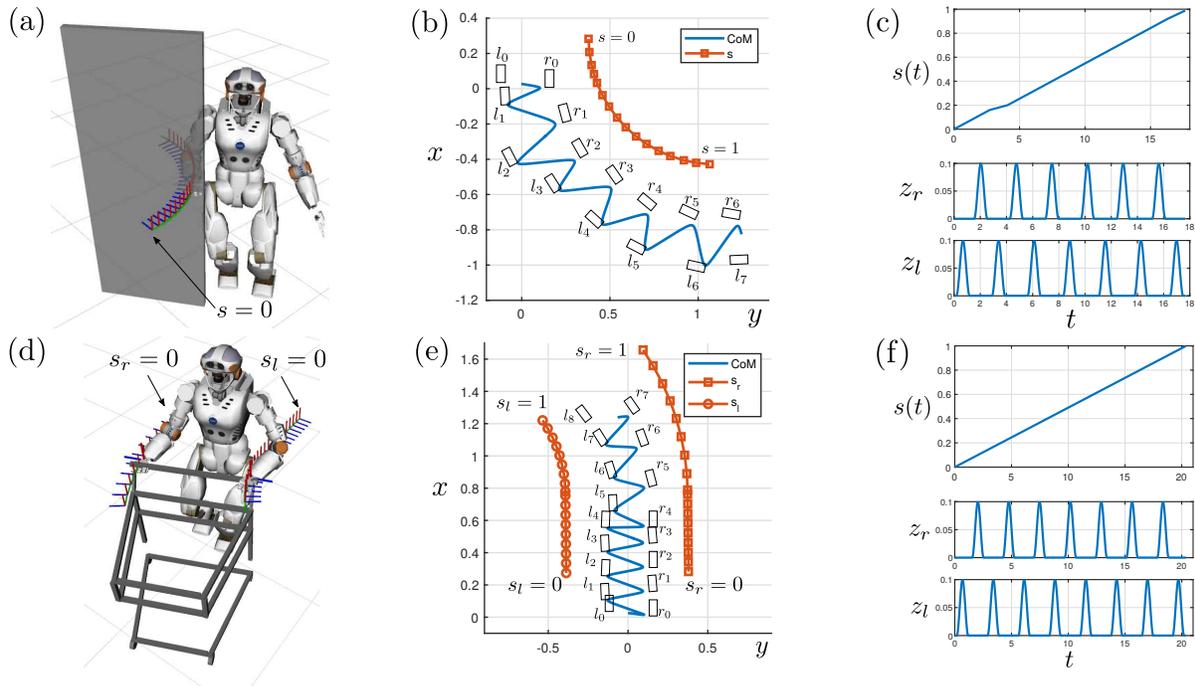}}
\caption{A 3D view of Valkyrie opening the door (a) and pushing a cart (d). (b) and (e) show a top-view of the center-of-mass trajectory(CoM), the manipulation end-effector trajectory $f(s)$ with $s=0$ and $s=1$ indicating the start and ending end-effector poses respectively, and the $i$-th left and right footsteps with $i=0$ being the starting stance location. (c) and (f) show the manipulation progression variable trajectory $s(t)$ as a function of time and a visualization of the footstep contact transitions using the z-height of the left and right footsteps.}
\label{fig:door_open_result}
\end{figure*}

\begin{table*}[ht]
\caption{ Planner Performance with and without the Classifier }
\centering
\label{table:planner_performance}
\begin{tabular}{|c||c|c|c|c|c|c|}
\hline
\multirow{2}{*}{\textbf{Planner Type}}  & \multicolumn{2}{c|}{\textbf{Time to Goal Vertex (secs)}} & \multicolumn{2}{c|}{\textbf{Reconstruction Time (secs)}} & \multicolumn{2}{c|}{\textbf{Total Planning Time (secs)}} \\
\cline{2-7}
 & Door Opening & Cart Pushing & Door Opening & Cart Pushing & Door Opening & Cart Pushing \\
\hline
With the Classifier         & \textbf{4.78s} & \textbf{3.32s} & 28.31s &  25.41s & \textbf{33.09s} &  \textbf{28.73s}\\
\hline
Without the Classifier  & 38.29s & 32.24s & 0.0 & 0.0 & 38.29s & 32.24s \\
\hline
\end{tabular}
\end{table*}

\subsubsection{Edge Transition Feasibility}
To increase efficiency, all neighbors are assumed to be feasible until the vertex is extracted from the prioritiy queue. When the assumed feasible vertex is extracted from the queue, edge validity is performed by testing if a feasible transition exists from $v_1$ to $v_2$. Without a classifier, this requires the use of solving the IK configuration trajectory between two vertices. With the classifier, if the feasibility score is greater than $0.5$, the transition is assumed to be feasible. If the edge between $v_1$ and $v_2$ is not feasible, then the next vertex in the priority queue is processed. As the classifier can make mistakes on feasible and unfeasible transitions, we reconstruct the full path with the IK when the goal vertex has been reached and only return the candidate plan if it converged. This reconstruction step is not needed if the classifier is not used.

The weighted A* is used as the planner \cite{ebendt2009weighted} for the graph search problem to produce sub-optimal but faster plans than A*. The following heuristic $h(v)$ with scalar weight $w_h$ brings $s$ to $1$ with
\begin{align}
\label{eq:astar_heuristic}
    h(v) = w_h\cdot(w_s (1-s)).
\end{align}
When $w_h = 1$, the solution of the planner is the optimal result produced by the A* as the heuristic is admissible \cite{norvig1995modernapproach} since Eq.~\ref{eq:astar_heuristic} will be equal to the first term of Eq~\ref{eq:edge_cost}. Similar to \cite{lin2019efficient}, we use an $\epsilon$-greedy strategy \cite{valenzano2014comparison} to aid escaping cul-de-sac scenarios by randomly evaluating a vertex in the priority queue with probability $\epsilon$ $(0 < \epsilon < 1)$. 

% \subsubsection{Classifier-based Edge Transition Evaluation}
% \label{sec:classifier-based-edge-transition-evaluation}

%where $(\cdot)_{\rm feet} = [(\cdot)^{\rm left}_{\rm foot}, (\cdot)^{\rm right}_{\rm foot}]^T \in \mathbf{R}^2$. When $\Delta s$ is non-zero, this indicates a progression of the manipulation variable $s$. Similarly, when the feet variables are non-zero this indicates that a footstep would occur. Naturally, both feet cannot be non-zero at the same time as the stance leg needs to be zero. To make the problem tractable, a finitely sized lattice is considered with a radius of $2m$ from the stance origin. As the \rt{manipulation function} $f(s)$ defines the path of the end-effector, there will be vertices with stance positions that are  kinematically too far. 

\section{RESULTS and CONCLUSIONS}
We provide two toy examples\footnote{\url{https://youtu.be/C4pfgatgYFE}} in which locomanipulation is achieved for a given end-effector task space trajectory. Fig. \ref{fig:door_open_result} shows a figure of Valkyrie opening a door and performing bimanual push of a cart. Table~\ref{table:edge_feasibility_check_performance} shows that the classifier evaluates edge transitions very efficiently. Table~\ref{table:planner_performance} shows that utilizing the locomanipulability classifier can find goal vertices faster, but the reconstruction step for confirming the full trajectory feasibility is a bottleneck. Still, a properly trained classifier can return results faster than without it, and a better implementation of the reconstruction step should decrease the overall planning time. 

%\rt{From empirical testing, we have found that the planner success relies on good initial conditions of the robot configuration at $s = 0$. That is, if the initial robot configuration is not at a high locomanipulability region, the planner spends significant amount of time escaping the low locomanipulability region before being able to find a feasible locomanipulation plan. Planner success also relies on the quality of the learned classifier.} 

To conclude, we have demonstrated a fast approach for finding locomanipulation plans by finding admissible manipulation trajectories in the constraint manifold. %We used a weighted A* planner to perform search on a lower dimensional problem of locomanipulation and used a neural network to quickly classify feasible contact transitions. 
While our approach produces kinodynamic plans, our method relies on the user or another high-level planner to provide end-effector plans which may not be a correct manipulation description for the object. While the full-body plans can be immediatley used on the robot by using its existing API as done previously in \cite{jorgensen2019thermal}, a robust implementation would require online replanning of hand trajectories (e.g \cite{arduengo2019versatile}) if deviations in forces or kinematic trajectories have been detected.
%Additionally, finds locomanipulation plans given end-effector paths. Extensions to include dynamics given the kinematic plans are possible for instance with Time Optimal Path Planning (TOPP) \rt{cite Hauser} or SQP \rt{cite hongkai} formulations. However, execution on the real robot is expected to be feasible provided that the disturbance introduced by the manipulation trajectories while handling objects are sufficiently rejected by the controller such that the locomotion tasks are still controllable.


%%%%%%%%%%%%%%%%%%%%%%%%%%%%%%%%%%%%%%%%%%%%%%%%%%%%%%%%%%%%%%%%%%%%%%%%%%%%%%%%

%%%%%%%%%%%%%%%%%%%%%%%%%%%%%%%%%%%%%%%%%%%%%%%%%%%%%%%%%%%%%%%%%%%%%%%%%%%%%%%%

%%%%%%%%%%%%%%%%%%%%%%%%%%%%%%%%%%%%%%%%%%%%%%%%%%%%%%%%%%%%%%%%%%%%%%%%%%%%%%%%
\section*{ACKNOWLEDGMENT}
We are grateful to Junhyeok Ahn of the Human Centered Robotics Lab for his C++ neural network code.
This work is supported by a NASA Space Technology Research Fellowship (NSTRF) grant \#NNX15AQ42H.

%%%%%%%%%%%%%%%%%%%%%%%%%%%%%%%%%%%%%%%%%%%%%%%%%%%%%%%%%%%%%%%%%%%%%%%%%%%%%%%%
\bibliographystyle{IEEEtran}
\bibliography{references}

\begin{thebibliography}{10}
\providecommand{\url}[1]{#1}
\csname url@rmstyle\endcsname
\providecommand{\newblock}{\relax}
\providecommand{\bibinfo}[2]{#2}
\providecommand\BIBentrySTDinterwordspacing{\spaceskip=0pt\relax}
\providecommand\BIBentryALTinterwordstretchfactor{4}
\providecommand\BIBentryALTinterwordspacing{\spaceskip=\fontdimen2\font plus
\BIBentryALTinterwordstretchfactor\fontdimen3\font minus
  \fontdimen4\font\relax}
\providecommand\BIBforeignlanguage[2]{{%
\expandafter\ifx\csname l@#1\endcsname\relax
\typeout{** WARNING: IEEEtran.bst: No hyphenation pattern has been}%
\typeout{** loaded for the language `#1'. Using the pattern for}%
\typeout{** the default language instead.}%
\else
\language=\csname l@#1\endcsname
\fi
#2}}

\bibitem{meeussen2010autonomous}
W.~Meeussen, M.~Wise, S.~Glaser, S.~Chitta, C.~McGann, P.~Mihelich,
  E.~Marder-Eppstein, M.~Muja, V.~Eruhimov, T.~Foote, \emph{et~al.},
  ``Autonomous door opening and plugging in with a personal robot,'' in
  \emph{2010 IEEE International Conference on Robotics and Automation}.\hskip
  1em plus 0.5em minus 0.4em\relax IEEE, 2010, pp. 729--736.

\bibitem{ruhr2012generalized}
T.~R{\"u}hr, J.~Sturm, D.~Pangercic, M.~Beetz, and D.~Cremers, ``A generalized
  framework for opening doors and drawers in kitchen environments,'' in
  \emph{2012 IEEE International Conference on Robotics and Automation}.\hskip
  1em plus 0.5em minus 0.4em\relax IEEE, 2012, pp. 3852--3858.

\bibitem{welschehold2017learning}
T.~Welschehold, C.~Dornhege, and W.~Burgard, ``Learning mobile manipulation
  actions from human demonstrations,'' in \emph{2017 IEEE/RSJ International
  Conference on Intelligent Robots and Systems (IROS)}.\hskip 1em plus 0.5em
  minus 0.4em\relax IEEE, 2017, pp. 3196--3201.

\bibitem{arduengo2019versatile}
M.~Arduengo, C.~Torras, and L.~Sentis, ``A versatile framework for robust and
  adaptive door operation with a mobile manipulator robot,'' \emph{arXiv
  preprint arXiv:1902.09051}, 2019.

\bibitem{posa2013direct}
M.~Posa and R.~Tedrake, ``Direct trajectory optimization of rigid body
  dynamical systems through contact,'' in \emph{Algorithmic foundations of
  robotics X}.\hskip 1em plus 0.5em minus 0.4em\relax Springer, 2013, pp.
  527--542.

\bibitem{dalibard2010manipulation}
S.~Dalibard, A.~Nakhaei, F.~Lamiraux, and J.-P. Laumond, ``Manipulation of
  documented objects by a walking humanoid robot,'' in \emph{10th IEEE-RAS
  International Conference on Humanoid Robots}, 2010, pp. pp--518.

\bibitem{mirabel2016constraint}
J.~Mirabel and F.~Lamiraux, ``Constraint graphs: Unifying task and motion
  planning for navigation and manipulation among movable obstacles,''
  \emph{hal-01281348}, 2016.

\bibitem{berenson2011task}
D.~Berenson, S.~Srinivasa, and J.~Kuffner, ``Task space regions: A framework
  for pose-constrained manipulation planning,'' \emph{The International Journal
  of Robotics Research}, vol.~30, no.~12, pp. 1435--1460, 2011.

\bibitem{ferrari2017humanoid}
P.~Ferrari, M.~Cognetti, and G.~Oriolo, ``Humanoid whole-body planning for
  loco-manipulation tasks,'' in \emph{2017 IEEE International Conference on
  Robotics and Automation (ICRA)}.\hskip 1em plus 0.5em minus 0.4em\relax IEEE,
  2017, pp. 4741--4746.

\bibitem{englsberger2015three}
J.~Englsberger, C.~Ott, and A.~Albu-Sch{\"a}ffer, ``Three-dimensional bipedal
  walking control based on divergent component of motion,'' \emph{IEEE
  Transactions on Robotics}, vol.~31, no.~2, pp. 355--368, 2015.

\bibitem{radford2015valkyrie}
N.~A. Radford, P.~Strawser, K.~Hambuchen, J.~S. Mehling, W.~K. Verdeyen, A.~S.
  Donnan, J.~Holley, J.~Sanchez, V.~Nguyen, L.~Bridgwater, \emph{et~al.},
  ``Valkyrie: Nasa's first bipedal humanoid robot,'' \emph{Journal of Field
  Robotics}, vol.~32, no.~3, pp. 397--419, 2015.

\bibitem{koolen2016design}
T.~Koolen, S.~Bertrand, G.~Thomas, T.~De~Boer, T.~Wu, J.~Smith, J.~Englsberger,
  and J.~Pratt, ``Design of a momentum-based control framework and application
  to the humanoid robot atlas,'' \emph{International Journal of Humanoid
  Robotics}, vol.~13, no.~01, p. 1650007, 2016.

\bibitem{borras2015whole}
J.~Borras and T.~Asfour, ``A whole-body pose taxonomy for loco-manipulation
  tasks,'' in \emph{2015 IEEE/RSJ International Conference on Intelligent
  Robots and Systems (IROS)}.\hskip 1em plus 0.5em minus 0.4em\relax IEEE,
  2015, pp. 1578--1585.

\bibitem{farnioli2016toward}
E.~Farnioli, M.~Gabiccini, and A.~Bicchi, ``Toward whole-body
  loco-manipulation: Experimental results on multi-contact interaction with the
  walk-man robot,'' in \emph{2016 IEEE/RSJ International Conference on
  Intelligent Robots and Systems (IROS)}.\hskip 1em plus 0.5em minus
  0.4em\relax IEEE, 2016, pp. 1372--1379.

\bibitem{kaiser2016towards}
P.~Kaiser, E.~E. Aksoy, M.~Grotz, and T.~Asfour, ``Towards a hierarchy of
  loco-manipulation affordances,'' in \emph{2016 IEEE/RSJ International
  Conference on Intelligent Robots and Systems (IROS)}.\hskip 1em plus 0.5em
  minus 0.4em\relax IEEE, 2016, pp. 2839--2846.

\bibitem{asfour2018dualities}
T.~Asfour, J.~Borr{\`a}s, C.~Mandery, P.~Kaiser, E.~E. Aksoy, and M.~Grotz,
  ``On the dualities between grasping and whole-body loco-manipulation tasks,''
  in \emph{Robotics Research}.\hskip 1em plus 0.5em minus 0.4em\relax Springer,
  2018, pp. 305--322.

\bibitem{settimi2016motion}
A.~Settimi, D.~Caporale, P.~Kryczka, M.~Ferrati, and L.~Pallottino, ``Motion
  primitive based random planning for loco-manipulation tasks,'' in \emph{2016
  IEEE-RAS 16th International Conference on Humanoid Robots (Humanoids)}.\hskip
  1em plus 0.5em minus 0.4em\relax IEEE, 2016, pp. 1059--1066.

\bibitem{hart2014affordance}
S.~Hart, P.~Dinh, and K.~A. Hambuchen, ``Affordance templates for shared robot
  control,'' in \emph{2014 AAAI Fall Symposium Series}, 2014.

\bibitem{bouyarmane2012humanoid}
K.~Bouyarmane and A.~Kheddar, ``Humanoid robot locomotion and manipulation step
  planning,'' \emph{Advanced Robotics}, vol.~26, no.~10, pp. 1099--1126, 2012.

\bibitem{bouyarmane2018non}
------, ``Non-decoupled locomotion and manipulation planning for
  low-dimensional systems,'' \emph{Journal of Intelligent \& Robotic Systems},
  vol.~91, no. 3-4, pp. 377--401, 2018.

\bibitem{dai2014whole}
H.~Dai, A.~Valenzuela, and R.~Tedrake, ``Whole-body motion planning with
  centroidal dynamics and full kinematics,'' in \emph{2014 IEEE-RAS
  International Conference on Humanoid Robots}.\hskip 1em plus 0.5em minus
  0.4em\relax IEEE, 2014, pp. 295--302.

\bibitem{kim2018control}
D.~Kim, S.~J. Jorgensen, H.~Hwang, and L.~Sentis, ``Control scheme and
  uncertainty considerations for dynamic balancing of passive-ankled bipeds and
  full humanoids,'' in \emph{2018 IEEE-RAS 18th International Conference on
  Humanoid Robots (Humanoids)}.\hskip 1em plus 0.5em minus 0.4em\relax IEEE,
  2018, pp. 1--9.

\bibitem{ponton2018time}
B.~Ponton, A.~Herzog, A.~Del~Prete, S.~Schaal, and L.~Righetti, ``On time
  optimization of centroidal momentum dynamics,'' in \emph{2018 IEEE
  International Conference on Robotics and Automation (ICRA)}.\hskip 1em plus
  0.5em minus 0.4em\relax IEEE, 2018, pp. 1--7.

\bibitem{zacharias2007capturing}
F.~Zacharias, C.~Borst, and G.~Hirzinger, ``Capturing robot workspace
  structure: representing robot capabilities,'' in \emph{2007 IEEE/RSJ
  International Conference on Intelligent Robots and Systems}.\hskip 1em plus
  0.5em minus 0.4em\relax IEEE, 2007, pp. 3229--3236.

\bibitem{shoemake1985animating}
K.~Shoemake, ``Animating rotation with quaternion curves,'' in \emph{ACM
  SIGGRAPH computer graphics}, vol.~19, no.~3.\hskip 1em plus 0.5em minus
  0.4em\relax ACM, 1985, pp. 245--254.

\bibitem{kim1995general}
M.-J. Kim, M.-S. Kim, and S.~Y. Shin, ``A general construction scheme for unit
  quaternion curves with simple high order derivatives,'' in \emph{SIGGRAPH},
  vol.~95, 1995, pp. 369--376.

\bibitem{lin2019efficient}
Y.-C. Lin, B.~Ponton, L.~Righetti, and D.~Berenson, ``Efficient humanoid
  contact planning using learned centroidal dynamics prediction,'' in
  \emph{2019 International Conference on Robotics and Automation (ICRA)}.\hskip
  1em plus 0.5em minus 0.4em\relax IEEE, 2019, pp. 5280--5286.

\bibitem{nair2010rectified}
V.~Nair and G.~E. Hinton, ``Rectified linear units improve restricted boltzmann
  machines,'' in \emph{Proceedings of the 27th international conference on
  machine learning (ICML-10)}, 2010, pp. 807--814.

\bibitem{chollet2015}
F.~Chollet, ``keras,'' \url{https://github.com/fchollet/keras}, 2015.

\bibitem{ebendt2009weighted}
R.~Ebendt and R.~Drechsler, ``Weighted a∗ search--unifying view and
  application,'' \emph{Artificial Intelligence}, vol. 173, no.~14, pp.
  1310--1342, 2009.

\bibitem{norvig1995modernapproach}
S.~J. Russell and P.~Norvig, \emph{Artificial Intelligence: A Modern
  Approach}.\hskip 1em plus 0.5em minus 0.4em\relax Prentice-Hall, 1995.

\bibitem{valenzano2014comparison}
R.~A. Valenzano, N.~R. Sturtevant, J.~Schaeffer, and F.~Xie, ``A comparison of
  knowledge-based gbfs enhancements and knowledge-free exploration,'' in
  \emph{Twenty-Fourth International Conference on Automated Planning and
  Scheduling}, 2014.

\bibitem{jorgensen2019thermal}
S.~J. Jorgensen, J.~Holley, F.~Mathis, J.~S. Mehling, and L.~Sentis, ``Thermal
  recovery of multi-limbed robots with electric actuators,'' \emph{IEEE
  Robotics and Automation Letters}, vol.~4, no.~2, pp. 1077--1084, 2019.

\end{thebibliography}

\end{document}